\documentclass[sigconf]{acmart}
\usepackage{balance}
\usepackage{soul}
\usepackage{booktabs} 
\usepackage{graphicx}
\usepackage{subfigure}
\usepackage{amsthm}
\usepackage{amsmath}
\usepackage{multirow}
\usepackage{enumerate}
\usepackage{amsfonts}
\usepackage{color}
\usepackage{url}
\usepackage{amsfonts}
\usepackage{tabularx}
\usepackage{diagbox}
\usepackage{longtable}
\usepackage{rotating}
\usepackage{slashbox}
\usepackage{caption}
\usepackage[ruled, linesnumbered]{algorithm2e}
\usepackage{bm}
\usepackage{float}
\usepackage{hyperref}

\def\model{EQUATE}
\AtBeginDocument{%
  }
\renewcommand\footnotetextcopyrightpermission[1]{} 
\AtBeginDocument{\pagestyle{plain}\fancyhead{}}

\begin{document}

\title{Data2Eqn: A Fine-tuning Framework for Foundation Knowledge Transfer in Generative Model Equation Learning}
\title{Data2Eqn: A Foundation Knowledge Transfer Framework for Generative Model Equation Learning}
\title{A Foundation Knowledge Transfer Framework for Generative Model Equation Distillation with Small Tasking Data}
\title{Domain-Aware Fine-Tuning for Symbolic Regression}
\title{Data-Efficient Symbolic Regression via Foundation Model Distillation}  
\author{Wangyang Ying}
\orcid{0009-0009-6196-0287}
\affiliation{%
  \institution{Arizona State University}
  \city{Tempe}
  \state{Arizona}
  \country{USA}
}\email{wangyang.ying@asu.edu}

\author{Jinghan Zhang}
\affiliation{%
  \institution{Clemson University}
  \city{Clemson}
  \state{South Carolina}
  \country{USA}}
\email{jinghaz@clemson.edu}

\author{Haoyue Bai}
\affiliation{%
  \institution{Arizona State University}
  \city{Tempe}
  \state{Arizona}
  \country{USA}
}\email{haoyuebai@asu.edu}

\author{Nanxu Gong}
\affiliation{%
  \institution{Arizona State University}
  \city{Tempe}
  \state{Arizona}
  \country{USA}
}\email{nanxugong@asu.edu}

\author{Xinyuan Wang}
\affiliation{%
  \institution{Arizona State University}
  \city{Tempe}
  \state{Arizona}
  \country{USA}
}\email{xwang735@asu.edu}

\author{Kunpeng Liu}
\affiliation{%
  \institution{Clemson University}
  \city{Clemson}
  \state{South Carolina}
  \country{USA}}
\email{kunpenl@clemson.edu}

\author{Chandan K. Reddy}
\affiliation{%
  \institution{Virginia Tech}
  \city{Blacksburg}
  \state{Virginia}
  \country{USA}}
\email{reddy@cs.vt.edu}

\author{Yanjie Fu}
\orcid{0000-0002-1767-8024}
\affiliation{%
 \institution{Arizona State University}
 \city{Tempe}
 \state{Arizona}
 \country{USA}}
\email{yanjie.fu@asu.edu}

\renewcommand{\shortauthors}{Ying et al.}

\begin{abstract}
Discovering interpretable mathematical equations from observed data (a.k.a. equation discovery or symbolic regression) is a cornerstone of scientific discovery, enabling transparent modeling of physical, biological, and economic systems. While foundation models pre-trained on large-scale equation datasets offer a promising starting point, they often suffer from negative transfer and poor generalization when applied to small, domain-specific datasets.
In this paper, we introduce EQUATE (Equation Generation via QUality-Aligned Transfer Embeddings), a data-efficient fine-tuning framework that adapts foundation models for symbolic equation discovery in low-data regimes via distillation. EQUATE combines symbolic-numeric alignment with evaluator-guided embedding optimization, enabling a principled embedding-search-generation paradigm. Our approach reformulates discrete equation search as a continuous optimization task in a shared embedding space, guided by data-equation fitness and simplicity. Experiments across three standard public benchmarks (Feynman, Strogatz, and black-box datasets) demonstrate that EQUATE consistently outperforms state-of-the-art baselines in both accuracy and robustness, while preserving low complexity and fast inference. These results highlight EQUATE as a practical and generalizable solution for data-efficient symbolic regression in foundation model distillation settings.
\end{abstract}

\maketitle

\vspace{-0.2cm}
\section{Introduction}
Many scientific and engineering applications require uncovering the underlying mathematical equations that describe observed data. This symbolic regression task aims to discover interpretable model equations that capture relationships within data without relying on predefined functional forms. Such equations support transparent modeling, inverse engineering, and low-cost discovery of underlying physical, biological, or economic laws~\cite{wang2019symbolic,angelis2023artificial,cranmer2020discovering,can2011comparison,claveria2017assessment,la2023flexible}.

Prior literature on model equation discovery is two-fold: 1) Genetic Programming (GP) based methods~\cite{GP,GP2,GP3,GP6,GP7}; 2) deep learning based methods~\cite{petersen2019deep,kim2020integration,zhang2023deep,d2022deep}. Unfortunately, GP methods suffer from high search complexity and thus are inefficient and slow; deep learning methods cannot apply to single small-scale task-specific datasets (i.e., perform poorly when large-scale data are not available).
Emerging foundation model techniques can learn generalizable knowledge and robust representations from massive datasets and become prevalent in model equation discovery. For example, a Transformer-based foundation equation discovery model was proposed in~\cite{biggio2021neural} via large-scale pre-training on generic datasets, followed by other foundation models~\cite{kamienny2022end,shojaee2023transformer,holt2023deep,valipour2021symbolicgpt}. The underlying idea of these foundation models is to encode a dataset into an embedding vector, then decode the embedding vector to the corresponding equation.
This motivates our study of data-efficient symbolic regression via foundation model distillation: How can we effectively distill general symbolic knowledge from foundation models into compact, task-specific equations using only small datasets?

\begin{figure}[t]
\centering
\includegraphics[width=\linewidth]{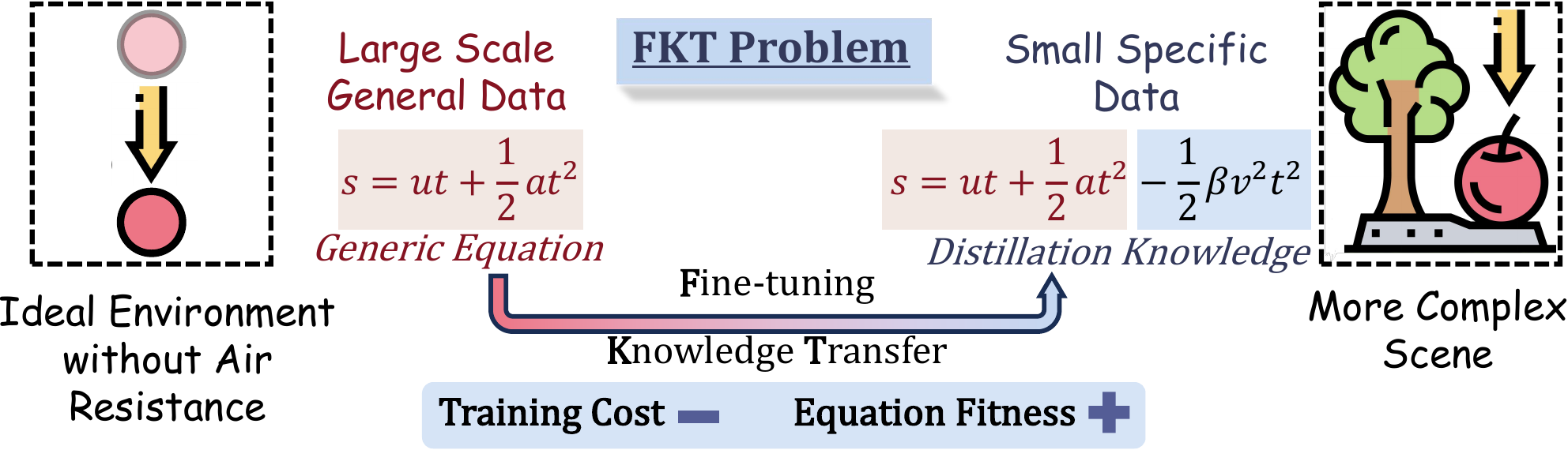}
\vspace{-0.5cm}
\caption{FKT transfers generic knowledge to specific domains through fine-tuning and knowledge distillation.}
\vspace{-0.5cm}
\label{motivation}
\end{figure}

We formalize this as the FKT problem: developing a dedicated \textbf{F}inetuning framework for foundation \textbf{K}nowledge \textbf{T}ransfer in small-data symbolic regression tasks. As illustrated in \textbf{Figure~\ref{motivation}}, this framework should preserve useful foundation model priors, tailor them to task-specific data, and avoid the cost of training from scratch (as shown in \textbf{Figure~\ref{motivation}}). 

There are two main challenges in solving FKT: 
\textbf{1) representation alignment of symbolic equation, numeric data, and data-equation fitness.}
Equation spaces (symbolic, discrete), data spaces (numeric, continuous), and data-equation fitness (pairwise, scored) are structurally incompatible. How can we learn an aligned embedding space that enables seamless transfer between these modalities?
\textbf{2) optimization steering for structured and meaningful task-specific search of equations.} 
The generation objectives of existing equation discovery foundation models are typically borrowed from textual generation pre-training, such as token-level cross-entropy loss. However, classic textual generation objectives tend to generate equations with high token-level similarity. 
This optimization direction fundamentally deviates from the desired directions of equation discovery: high equation discovery fitness and equation simplicity.
The optimization challenge aims to answer: How can we develop an optimization feedback mechanism to allow for a more structured and meaningful search of best-fitting equations under small tasking data?

\textbf{Our insight: integrating numeric-symbolic alignment and evaluator-guided optimization for transfer fine-tuning on small tasking data.} 
Our AI task is to develop a foundation knowledge transfer fine-tuning framework for model equation discovery from small tasking data. 
We found that an embedding-search-generation framework opens three new perspectives to augment equation discovery fine-tuning with small domain tasking data: 
1) The embedding step empowers us to redesign the encoder's neural structure (e.g.,  attentive fusion with pre-trained transformer encoder) to compress not just numeric data patterns but also symbolic equation structures to a symbol-numeric aligned embedding space of data-equation pairs, while still exploiting foundation model knowledge. Such an alignment strategy exhibits better representation expressiveness and truthfulness than the methods that only compress numerical data or directly fine-tune existing foundation equation discovery models. 
2) The embedding step creates a continuous space to represent data-equation fitness, thus reformulating discrete equation search into differentiable optimization. The gradient-based equation search step empowers us to redesign a specialized domain loss (i.e., trade-off between data-equation fitness and equation simplicity) for precision fine-tuning with small-tasking data. 
Interestingly, by adding and learning an evaluator to predict data-equation fitness and equation simplicity during fine-tuning, we find that the learned evaluator can provide task-specific gradient guidance (feedback) to steer and tailor equation search to domain data, and avoid misalignment when transferring foundation general knowledge to specific tasks. 
3) In our task setting, we are only given a single small task-specific dataset to be distilled into an equation.  We demonstrate that the data-equation fitness gap between the foundation model-generated equation and the best-fitting equation can serve as fine-tuning supervision signals. 

\textbf{Summary of technical solution.} We propose Equation Generation via Quality-Aligned Transfer Embeddings (\textbf{EQUATE}), which includes four steps: 
1) Training Set Preparation: We leverage a pre-trained foundation model to generate imperfect yet relevant candidate equations on the task-specific dataset. These equations are applied to sampled data subsets to create training instances labeled by their data-equation fitness scores.
2) Embedding Space Construction: We design a neural architecture comprising a partially frozen transformer encoder for numerical data, an LSTM encoder for symbolic equations, an attention-based fusion module, an evaluator, and a partially frozen decoder. This setup learns a shared symbolic-numeric embedding space that captures both structural and semantic fitness.
3) Fitness-Guided Search: We perform gradient-based optimization in the embedding space, using evaluator feedback to guide the search toward embeddings corresponding to high-fidelity equations.
4) Equation Generation: The optimized embedding is decoded via the autoregressive decoder to produce the final equation. The decoder is fine-tuned to align with domain-specific data while retaining prior knowledge from the foundation model.

\textbf{Our contributions.}
This paper introduces EQUATE, a fine-tuning framework that adapts foundation models for symbolic regression in small, domain-specific settings. Our key contributions are:
\ul{1) Fine-Tuning Framework for Symbolic Regression:} We propose a novel fine-tuning framework that adapts pretrained foundation models for symbolic equation generation to specific domains using only small task-specific datasets. This directly addresses the challenge of negative transfer commonly observed during foundation model adaptation.
\ul{2) Symbolic-Numeric Alignment for Knowledge Integration:} We develop a dual-encoder architecture that jointly embeds symbolic equations and numerical data into a shared representation space. This alignment allows the model to integrate domain-specific symbolic priors with observed numerical patterns, improving generalization and interpretability.
\ul{3) Evaluator-Guided Embedding Space Optimization:} We introduce a differentiable optimization strategy in embedding space, guided by an evaluator trained to capture both data-equation fitness and symbolic simplicity. This approach goes beyond standard token-level decoding, enabling structured, domain-aware search for optimal symbolic forms.
\ul{4) Fitness-Driven Symbolic Generation:} Unlike prior works that rely solely on token likelihoods, EQUATE incorporates fitness-based supervision during both training and inference. This results in more accurate and compact symbolic expressions, particularly under low-resource conditions.
We evaluate EQUATE across three public benchmarks (Feynman, Strogatz, and black-box datasets), demonstrating superior performance in accuracy, robustness to noise, equation simplicity, and inference efficiency compared to strong baselines.
\vspace{-0.4cm}
\section{Background and Our Equation Discovery Task}

\noindent\textbf{Equation Discovery Problem (a.k.a. Symbolic Regression).} 
The equation discovery problem is to find a solution that automatically derives a model equation from data.  
Formally, given a data set $\mathcal{D} = (\mathbf{X}, y)$, where $\mathbf{X}$ represents input observations and $y$ denotes the corresponding labels, the objective is to discover an optimal mathematical expression $f$ automatically. This expression should describe the relationship between $\mathbf{X}$ and $y$ in an interpretable manner, without requiring prior specification of the model structure.

\noindent\textbf{Transformer-based Foundation Equation Discovery Model.} 
A Transformer-based foundation equation discovery model~\cite{biggio2021neural} is pre-trained on various datasets consisting of input-output pairs $(\mathbf{X}, y)$ and corresponding symbolic equations that capture the relationship between features and gold labels. These equations are represented as token sequences in the prefix expression format. 
The idea of this foundation model is to encode each dataset into an embedding vector and then decode the embedding vector to the corresponding equation.  
However, without fine-tuning on a domain-specific dataset, the foundation model does not include specialized knowledge relevant to specific domains (e.g. medicine, law, or finance), does not reflect the distribution of the target-specific dataset, and may include conflicting generic knowledge. This phenomenon, known as negative transfer, limits the model’s performance in specialized settings. 

\begin{figure*}[t]
\centering
\includegraphics[width=\linewidth]{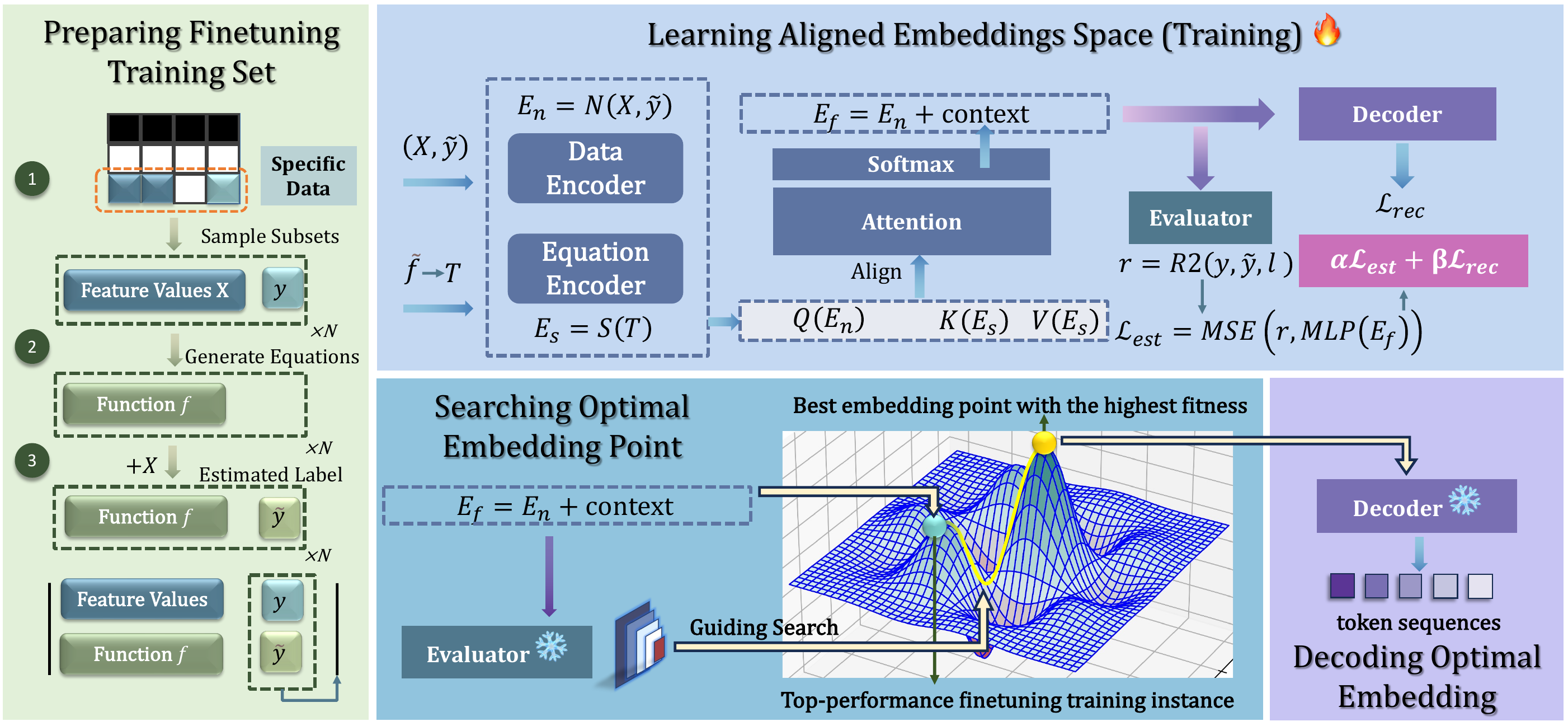}
\vspace{-0.5cm}
\caption{
Overview of the \model\ framework: a fine-tuning architecture that embeds data and symbolic equations, aligns them, and uses evaluator-guided optimization to distill foundation model knowledge for equation discovery on small datasets.
}

\vspace{-0.3cm}
\label{framework}
\end{figure*}

\noindent\textbf{Equation Generation via Quality-Aligned Transfer Embeddings (EQUATE).} 
Given a specific dataset and a foundation model pre-trained on equation discovery tasks, the objective of this study is to develop a fine-tuning framework that effectively transfers, adapts, and refines generic knowledge from the foundation model to produce symbolic equations tailored to the target domain.
This process bridges the gap between general-purpose representations and specialized domain knowledge by aligning with the target data distribution and mitigating the risk of negative transfer. The framework includes the following components:

\ul{\textit{1) Data:}} A Dedicated training dataset is needed to fine-tune the generative equation discovery model for a specific dataset. This involves a strategic extraction procedure to collect relevant training instances composed of data, symbolic equations, and fitness scores.
Formally,  given a specific dataset $\mathcal{D} = (\mathbf{X}, y)$, we prepare various data subsets $(\{(\mathbf{X}_1, ..., \mathbf{X}_n\}$, and various model equations that consist of symbolic expressions $\{\Tilde{f}_1, ..., \Tilde{f}_n\}$. Let $\Tilde{y}_i$ be the equation-estimated labels, we can denote our training set as $\{(\mathbf{X}_1, y_1, \Tilde{f}_1, \Tilde{y}_1),..., (\mathbf{X}_n, y_n, \Tilde{f}_n, \Tilde{y}_n\})$, whereby $\Tilde{y}_i = \Tilde{f}_i(\mathbf{X}_i)$, and $n$ is the number of data subsets. 

\ul{\textit{2) Encoder:}} 
The objective of the encoder is to learn expressive representations of data, equations, and fitness by (i) preserving general knowledge,  (ii) incorporating domain-specific information from the specific dataset, and (iii) aligning symbolic and numerical modalities. In this way, we emphasize features relevant to the specific dataset in a finetuned embedding space. 
Formally, let $\mathcal{N}$ be the data encoder of the pre-trained foundation model~\cite{kamienny2022end},  $\mathcal{S}$ is the equation encoder implemented by LSTM, $\circ$ is an abstract fusion and alignment operation, the embedding of data and equation is given by:
\begin{equation}
    E_f = \mathcal{N}(\mathbf{X}_i, \Tilde{y}_i) \circ \mathcal{S}(\Tilde{f}_i),
\end{equation}

\ul{\textit{3) Decoder:}} 
The objective of the decoder is to transform the latent representation of numeric, symbolic, and fitness information into the desired model equations that are interpretable and meaningful within the context of the specific dataset.  
Formally, let $\mathcal{I}$ be the fitness between gold labels (i.e., $y_i$) and function-estimated labels (i.e., $\Tilde{y}_i$), 
Our goal is to identify and decode the optimal embedding point of the highest data-equation fitness into the best-fitting equation $f^*$, given by:
\begin{equation}
    f^* = \arg\max_{E_f \in \mathcal{E}} \mathcal{I}(\psi(E_f)),
\end{equation}
where $\psi$ is the pre-trained decoder, and $\mathcal{E}$ is the embedding space. 
\section{Foundation Knowledge Transfer Finetuning for Equation Discovery}
\subsection{Overview}
Given a dataset and a foundation model for equation discovery, our objective is to leverage the knowledge of the foundation model and design a fine-tuning framework for learning equations from data. 
\textbf{Figure~\ref{framework}} illustrates our framework includes four steps:
1) preparing a training set of data subsets, equations, and fitness; 
2) learning the embedding space of data-equation pairs; 
3) identifying the optimal embedding; 4) generation of best-fit equations. 
In Step 1 (training set preparation), given a data table (features, gold labels), we prepare a training set that includes data, equations, and fitness for fine-tuning. In particular, we iterate the following process: firstly, sample a subset of instances from the data table; then, use the pre-trained foundation model to generate a relevant but imperfect model equation given the instance subset; and later, use the generated equation to predict equation-estimated labels for the subset. The fitness between a data subset and a model equation is measured by the gap between gold labels and equation-estimated labels. 
This iterative process generates quadruples, comprising the feature value matrix of an instance subset, a vector of gold labels, an equation, and a vector of equation-predicted labels, which form the training set. 
In Step 2 (embedding), we develop a deep representation model that includes a data encoder, equation encoder, evaluator, and equation decoder to learn the embedding space of data-equation pairs. In particular, the data encoder converts data to a vector; the equation encoder converts an equation to a vector; and the evaluator assesses the fitness between a data subset and an equation via regression over their fused embeddings. The objective is to learn parameters by minimizing the reconstruction loss between the decoded equation and the given equation and maximizing the accuracy of the evaluator. 
In Step 3 (search), after learning the representation embedding space of data-equation pairs for the given dataset, we aim to search for the best embedding point with the highest data-equation fitness, evaluated by the evaluator, in such space. 
In Step 4 (generation), we leverage the fine-tuned equation decoder to decode the best embedding point to the equation that best fits the data. 

\subsection{Preparing Finetuning Training Set: Data, Equation, and Fitness}
\noindent{\bf Why a dedicated fine-tuning training set in a dedicated format?}
Given a specific dataset and an equation discovery foundation model, to fine-tune a deep model to generate the best-fitting equation for the specific dataset, we need to construct a dedicated training set as fine-tuning supervision signals to learn the deep model via fine-tuning. 
To provide supervision for fine-tuning, such a training set should consist of several instances, each of which includes: Input data, Candidate equation, and Fitness score indicating how well the equation fits the data (objective to maximize). We highlight two core insights:
1) Since in the fine-tuning stage we only have one given target dataset, to construct many training instances, we can sample various subsets, as surrogates of the specific dataset, and the target from the specific dataset as inputs of \model. 
2) To simulate both good and bad fits, we leverage the foundation model to generate a range of imperfect yet relevant equations. The fitness is computed as the discrepancy between the gold and predicted labels. 
We develop a three-step method to prepare the fine-tuning training set:

\noindent\textbf{Step 1: prepare subsets of the specific dataset.} We use sampling algorithms to sample many instance subsets of the specific dataset, including both features and labels, as surrogates of the dataset. We now prepare many \{subset feature values, subset gold labels\} pairs. 

\noindent\textbf{Step 2: prepare relevant but imperfect equations.} We use the foundation equation discovery model to generate an equation for each data subset prepared in Step 1. We now prepare many \{subset feature values, subset gold labels, equation\} triplets. 

\noindent\textbf{Step 3: prepare equation-estimated labels.} For each triplet in Step 2, we apply the equation back to the subset feature values, to compute a list of equation-estimated labels for the subset. We now prepare many \{subset feature values, subset gold labels, equation, equation-estimated labels \} quadruples as the fine-tuning training set. By calculating the gap between subset gold labels and equation-estimated labels, we can quantify the fitness between an equation and a subset.  

\subsection{Learning Numeric and Symbolic Aligned Embeddings of Data and Equation Pairs}
\noindent{\bf Why a numeric and symbolic aligned embedding space?}
After preparing the fine-tuning training set of \{data feature values, data gold labels, equation, equation-estimated labels \}, a widely-used method is to train a deep model to encode data (a numeric space) to embedding and decode the embedding to equation (a symbolic space). 
Inspired by the text2image methods that embed both texts and images to ensure shared semantic space alignment, we propose to see a data-equation pair as a representation unit and learn a shared embedding space that aligns both numeric and symbolic meanings. There are two benefits: 1) \textbf{alignment:} It can improve alignment between numeric space (inputs: data) and symbolic space (outputs: tokens), enhance generalization by capturing deeper, more abstract, meaningful features that are invariant to noise and irrelevant details, moreover act as a regularizer to prevent overfitting.
2) \textbf{search:} In this aligned embedding space, each embedding vector represents a data-equation pair. The fitness of such pairs, if quantifiable by an evaluator, can be interpreted as the ``joint likelihood''  between data and equation. Finding the best equation for data can be solved like  ``joint likelihood maximization'', by gradient descent to search for the embedding of the highest fitness. 

\noindent\textbf{Leveraging numeric and symbolic alignment, attentive fusion, and data-equation fitness.}
We develop a deep embedding architecture that includes a data encoder for data embedding, an equation encoder for equation embedding, an attentive fusion for aligning data and equation embeddings, a decoder for equation generation, and an evaluator to measure data-equation fitness and enable gradient-steer search of the best-fitting equation embedding. 

\noindent\textbf{Step 1: Data embedding via data encoder.} 
We utilize the pre-trained encoder of the equation discovery foundation model~\cite{kamienny2022end} as our data encoder. 
This encoder embeds each training instance (i.e., subset feature values and equation-estimated labels) of the fine-tuning training set into a token sequence.
We then leverage a transformer to encode the token sequence into a fixed-length vector. 
Formally,  given the numeric observation of a training instance $(\mathbf{X}, \Tilde{y})$, where $\mathbf{X}$ is a subset of feature values, $\Tilde{y}$ is the equation-estimated label calculated by the equation $\Tilde{f}$. 
The data encoder $\mathcal{N}$ outputs a data embedding: $E_n = \mathcal{N}(\mathbf{X}, \Tilde{y})$.

\noindent\textbf{Step 2: Equation embedding via equation encoder.} 
We develop an LSTM-based equation encoder to learn the embedding of the equation corresponding to a numerical observation.  
Formally, given an equation $\Tilde{f}$, we exploit the same symbolic format of equations, defined by the equation discovery foundation model~\cite{kamienny2022end}, to convert $\Tilde{f}$ into a prefix expression. 
We then tokenize the prefix expression of an equation as $T = (t_1, t_2, ..., t_M)$, where $M$ is the number of tokens. The equation encoder outputs an equation embedding: $E_s = \mathcal{S}(T)$, where $\mathcal{S}$ is the equation encoder.

\noindent\textbf{Step 3: Attentive data-equation embedding fusion for numerical and symbolic alignment.} 
To align data and equation embeddings (denoted by $E_n$ and $E_s$), we employ an attention mechanism to empower the model to focus on the most relevant parts of the equation embedding with the data embedding. 
In particular, we regard the data embedding as the query $Q(E_n)$,  the equation embedding as both the key $K(E_s)$ and the value $V(E_s)$. 
Then, the attention scores are calculated as the scaled dot product of the query and the transposed key: $scores = Q(E_n)K(E_s)^\top/\sqrt{d_k}$, where $d_k$ is the dimensionality of the embeddings, and scaling by $\sqrt{d_k}$ helps stabilize the gradients during training. 
After that, the attention scores are normalized by the softmax function to produce attention weights: $W = softmax(scores)$. 
Later, the context vector is computed as the weighted sum of the value embeddings: $context = W \cdot V(E_s)$. 
Finally, we fuse the two vectors with residual connection: $E_f = E_n + context$. 
This idea allows the model to adapt to the dynamic relationships between data and equation embeddings through the attention weights, providing a more aligned representation of both numerical and symbolic spaces.

\noindent\textbf{Step 4: Evaluator to quantify fitness of data-equation embedding.} 
Aside from preserving both data and equation, it is essential for data-equation embedding to encode the fitness between data and equation into embedding and equation complexity. 
Our idea is to develop an evaluator to enforce that both fitness and complexity are predictable by the evaluator with data-equation embedding. 

To model fitness and complexity, we quantify the fitness or gap between equation-estimated labels $\Tilde{y}$, calculated by the equation $\Tilde{f}$, and the gold labels $y$, provided by the fine-tuning training set, by the R2-score between $\Tilde{y}$ and $y$.
\begin{equation}
    r = (1-\frac{\Sigma_{i=1}^n(\Tilde{y}_i-y_i)^2}{\Sigma_{i=1}^n(\Tilde{y}_i-\bar{y})^2}) + \lambda \cdot exp(-\frac{l(\Tilde{f})}{L}),
\end{equation}
where $l$ is the equation complexity represented by the length of $f$ in prefix expression, $L$ is the maximum sequence length, $\lambda$ is a hyperparameter to balance both fitness and complexity, $\bar{y}$ is the average of the true labels.
To construct an evaluator, we introduce a simple MLP following the attention layer to estimate $r$. Its loss function is denoted by: $\mathcal{L}_{est} = MSE(r, MLP(E_f))$ that provides gradient direction information to guide the search for the optimal point in the embedding space.

\noindent\textbf{Step 5: Decoder to generate corresponding equation.} 
The decoder is to reconstruct the tokenized equation $T=(t_1, t_2, ..., t_M)$. 
To transfer foundation knowledge, we exploit the pre-trained decoder of the foundation model. The decoder predicts the token distribution and generates the tokenized equation by taking the fused embedding $E_f$ as input. 

\noindent\textbf{Step 6: Objective function.}
During the autoregressive decoding process, assuming that the partially decoded sequence is $[t_1, t_2, ..., t_{j-1}]$, the decoder predicts the next token $t_j$.  The probability of generating the $j$-th token is given by:
\begin{equation}
    P_\mathcal{S}(t_j|E_f, [t_1, t_2, ..., t_{j-1}]) = \frac{exp(m_j)}{\Sigma_M exp(m)},
\end{equation}
where $m_j$ is the $j$-th output of the softmax layer. 
To optimize the reconstruction, we minimize the negative log-likelihood of generating sequential tokens to minimize the reconstruction loss between the predicted and ground truth tokens:
\begin{equation}
    \mathcal{L}_{rec} = -\Sigma_{j=1}^M log P_\mathcal{S}(t_j|E_f, [t_1, t_2, ..., t_{j-1}]).
\end{equation}
Moreover, we trade off the two losses to form a joint training loss: $\mathcal{L} = \alpha \cdot \mathcal{L}_{est} + \beta \cdot \mathcal{L}_{rec}$, where $\alpha$ and $\beta$ are the hyperparameters to balance the influence of the two loss functions. 

\subsection{Searching The Embedding Point of The Highest Data-Equation Fitness}
\textbf{Why search for the highest fitness embedding point?} After learning a continuous embedding space of data-equation pairs, a data-equation pair is represented by an embedding vector, along with a fitness score. 
Since the sample equations in the fine-tuning training instances are relevant but imperfect, we believe there exists the best embedding point with the highest fitness between the data and the equation described in this embedding space. 
With such continuous space and a fitness evaluator as directed guidance, we can find the best-fitting equation embedding to decode, so as to generate the optimal or near-optimal equation. 

\noindent\textbf{Leveraging continuous embedding space and gradient feedback of the evaluator to infer a search path toward the best embedding point.} In the embedding phrase, we evaluate each data pair using an evaluator to assess the fitness between equation-estimated labels and gold labels, as well as its complexity. 
During the inference phase, we search in the embedding space to identify the optimal data-equation representation.  
Our idea is to select the top-performance fine-tuning training instances as initial points to find better embedding points. In particular, we feed the subset feature values and corresponding equation-estimated label pair into the well-trained data encoder, and the equation into the equation encoders, to obtain a fused data-equation embedding $E_f$. 
Then, we use a gradient-ascent method to move the embedding with $\eta$ steps along the direction maximizing the accuracy $r$, denoted as $E_f^+ = E_f + \eta \cdot \frac{\partial{\theta}}{\partial{E_f}}$, where $\theta$ is the function notation of the evaluator.

\subsection{Decoding Optimal Embedding for Best Fitting Equation Generation}

Once we identify the optimal data-equation embedding point  $E_f^+$, we feed the optimal embedding point $E_f^+$ into the well-trained decoder to generate the equation token sequence in an autoregressive manner.  We iteratively generate the possible tokens until finding the end token (i.e., <EOS>) and transform this token sequence into an equation based on the predefined rules.

\subsection{Implementation Details of Foundation Knowledge Transfer and Finetuning}
The transformer-based equation discovery pre-trained model~\cite{kamienny2022end} offers foundational knowledge by modeling the generative relationships between diverse numerical data and massive symbolic equations. 
We leverage the foundation model in the fine-tuning pipeline as follows:

\noindent\textbf{1) Knowledge transfer:} 
Our data encoder and equation decoder of our fine-tuning pipeline share the same pre-trained weights as the data encoder and equation decoder of the foundation model. In this way, we transfer the general equation discovery knowledge in the foundation model to our fine-tuning pipeline, and avoid the risk of overfitting to limited task-specific data.

\noindent\textbf{2) Efficient fine-tuning:} 
During fine-tuning our embedding-search-generation pipeline, the weights of the data encoder and most of the decoder are frozen, while only the equation encoder and the last transformer layer of the decoder are retrained. 
This training strategy reduces the computational cost and the risk of catastrophic forgetting while enabling effective domain-specific adaptation.
\section{Experimental Results}

\setlength{\tabcolsep}{2.7mm}{
\begin{table*}[t]
\centering
\small
\caption{Overall performance compared with the foundation model E2E~\cite{kamienny2022end}. $R^2 > 0.99$ indicates the proportion of discovered equations on the dataset where the test set achieves $R^2 > 0.99$. $R^2$ represents the average test set performance across all discovered equations, while Complexity denotes the average complexity of the discovered equations.}
\vspace{-0.2cm}
\begin{tabular}{@{}cccclccclcc@{}}
\toprule\toprule
\multirow{2}{*}{Model} & \multicolumn{3}{c}{Feynman}                                                &  & \multicolumn{3}{c}{Strogatz}                                               &  & \multicolumn{2}{c}{Black-box}     \\ \cmidrule(lr){2-4} \cmidrule(lr){6-8} \cmidrule(l){10-11} 
                       & $\uparrow R^2$ > 0.99 & $\uparrow R^2$ & $\downarrow$ Complexity &  & $\uparrow R^2$ > 0.99 & $\uparrow R^2$ & $\downarrow$ Complexity &  & $\uparrow R^2$ & $\downarrow$ Complexity \\ \midrule
E2E-Beam               & 0.798                                  & 0.9727               & 54.44      &  & 0.286                                  & 0.7591               & 50.93      &  & 0.6968               & 81.72      \\
E2E-Sampling           & 0.815                                  & 0.9730               & \textbf{53.61}      &  & 0.357                                  & 0.8156               & 52.29      &  & 0.7225               & 83.18      \\ \midrule
\model-Beam               & 0.857                                  & 0.9765               & 54.78      &  & 0.714                                  & 0.9766               & 51.34      &  & 0.7789               & 82.56      \\
\model-Sampling                  & \textbf{0.874}                                  & \textbf{0.9833}               & 57.76      &  & \textbf{0.786}                                      & \textbf{0.9818}                   & \textbf{50.11}          &  & \textbf{0.7985}                    & \textbf{79.81}          \\ \bottomrule\bottomrule
\end{tabular}
\vspace{-0.3cm}
\label{exp:overall_performance}
\end{table*}}
\begin{figure*}[t]
  \centering
  \subfigure[\small{Feynman}]{
    \includegraphics[width=0.487\textwidth]{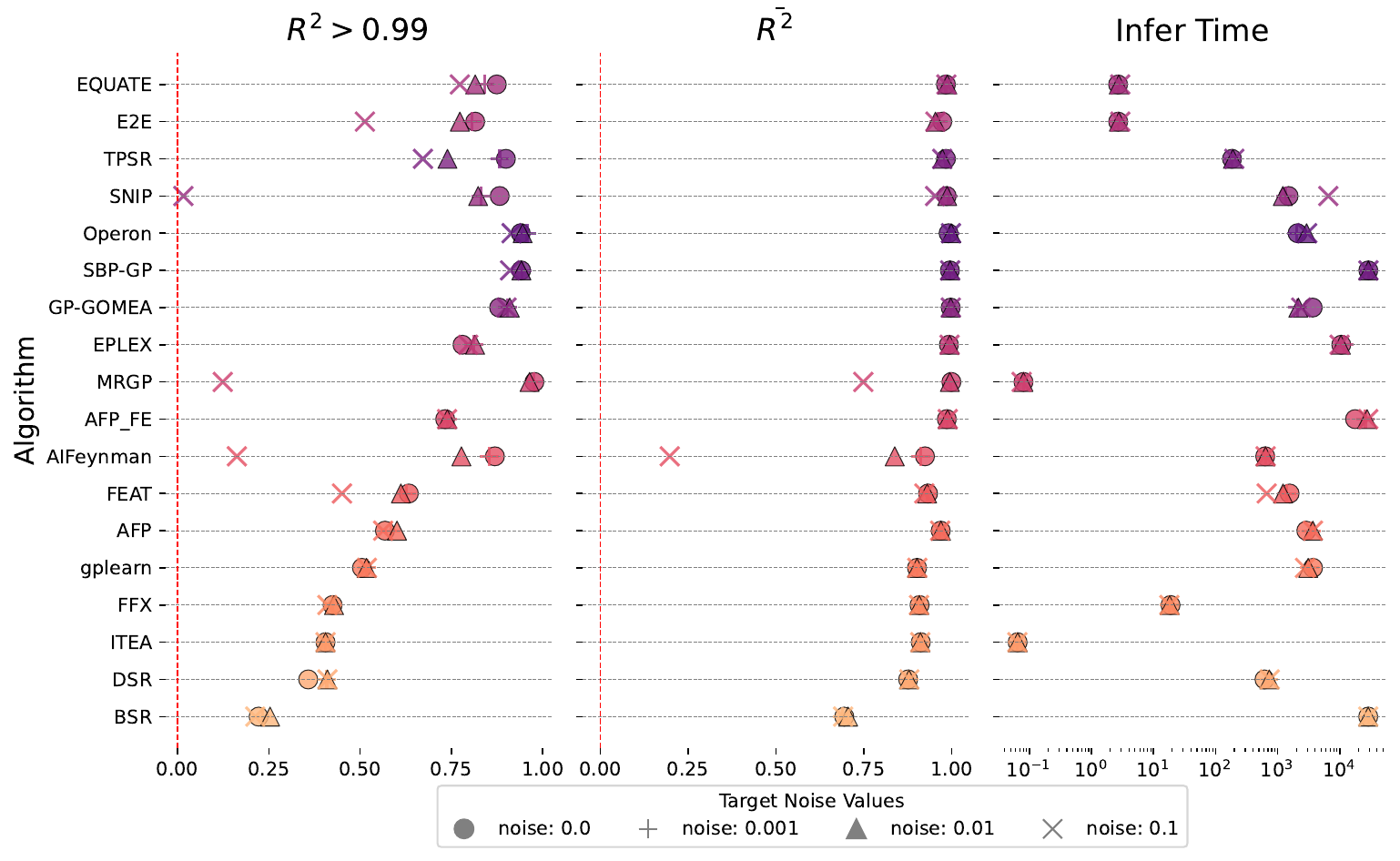}
  }
  \hfill
  \subfigure[\small{Strogatz}]{
    \includegraphics[width=0.487\textwidth]{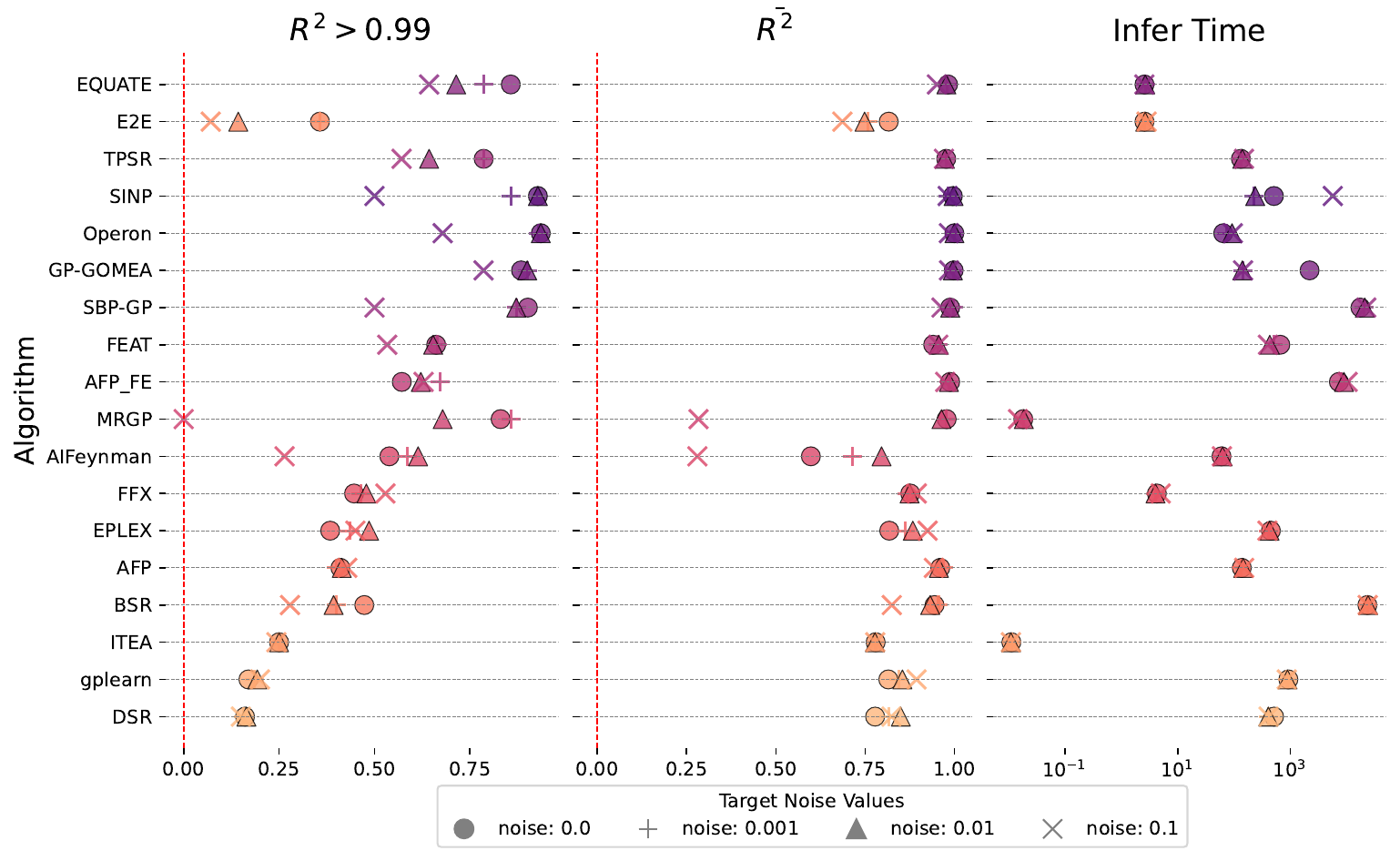}
  }
  \vspace{-0.4cm}
  \caption{Performance comparison of {\model} and all SRBench algorithms for Feynman and Strogatz datasets.}
  \vspace{-0.3cm}
  \label{fig:3in1}
\end{figure*}

\subsection{Experimental Settings}
\noindent\textbf{Datasets.} We used three real-world datasets:
\textbf{1) Feynman:} The Feynman dataset comprises 119 equations~\cite{udrescu2020aifeynman20paretooptimal}. For each equation, the input pairs $(X,y)$ are sampled by Penn Machine Learning Benchmark (PMLB)~\cite{olson2017pmlblargebenchmarksuite} and are utilized in SRBench~\cite{lacava2021contemporarysymbolicregressionmethods}. Each equation has an input dimensionality of less than 10. To construct the fine-tuning dataset, we sample B bags of 200 input points from the training data of each equation. This setup aligns with the pretraining data of the Transformer backbone~\cite{kamienny2022end}, which is trained on large datasets with input points $\leq$200.
\textbf{2) Strogatz:} The Strogatz dataset is constructed based on nonlinear dynamics and consists of 14 equations~\cite{la2016inference}. It is widely used in symbolic regression research. Each equation in the dataset corresponds to an input dimensionality of 2.
\textbf{3) Black-box:} The black-box dataset is sourced from PMLB and has been studied in multiple symbolic regression baselines. Since the pretraining of the foundation model strictly limits the input feature dimensionality to 10 or less, we select only those black-box datasets that meet this criterion. Finally, the dataset consists of a total of 57 equations.

\noindent\textbf{Evaluation.} 
1) \textbf{$\bm{R^2}$ score:} $R^2 = 1-\frac{\Sigma_{i=1}^n(\Tilde{y}_i-y_i)^2}{\Sigma_{i=1}^n(\Tilde{y}_i-\bar{y})^2}$, where $R^2$ measure the fitness accuracy of the generated equations. 
2) \textbf{$\bm{R^2 > 0.99}$:} For a given domain-specific dataset (e.g. Feynman), $R^2 > 0.99$ denotes the proportion of equations achieving an $R^2$ score larger than 0.99. For example, if 80 out of 119 equations in the Feynman datasets reach this threshold, then $R^2 > 0.99 = \frac{80}{119} = 0.672$. 
3) \textbf{Data splitting \& testing:} We split the datapoints of each equation into training and testing datapoints at a ratio of 75\% and 25\%, respectively. 
The fine-tuning datasets are sampled from the training datapoints, while the test datapoints are used to evaluate model performance. 
For each test function, we input 10 data matrices to obtain 10 initial embeddings. Each embedding is updated using gradients for up to 20 steps. We limit the number of candidate equations to 100. The search stops once an equation with $R^2 > 0.99$ is found or the maximum number of candidates reaches 100.

\noindent\textbf{Baselines.}
We compare {\model} with the backbone E2E and various SRBench algorithms to evaluate our proposed method.
1) End-to-end symbolic regression with transformers (E2E)~\cite{kamienny2022end};
2) Operon~\cite{GP};
3) Semantic Backpropagation GP (SBP-GP)~\cite{SBP-GP}; 
4) Transformer-based Planning for Symbolic Regression (TPSR)~\cite{shojaee2023transformer};
5) GP-based Gene-pool Optimal Mixing Evolutionary Algorithm (GP-GOMEA)~\cite{GP-GOMEA};
6) Epsilon-Lexicase Selection (EPLEX)~\cite{EPLEX};
7) Multiple Regression GP (MRGP)~\cite{MRGP}; 
8) Age-Fitness Pareto Optimization (AFP)~\cite{AFP};
9) Age-Fitness Pareto Optimization with Co-evolved Fitness Predictors 
 (AFP\_FE)~\cite{AFP}; 
10) AIFeynman~\cite{udrescu2020aifeynman20paretooptimal}; 
11) Feature Engineering Automation Tool (FEAT)~\cite{FEAT};
12) gplearn;
13) Fast Function Extraction (FFX)~\cite{FFX};
14) Interaction-Transformation Evolutionary Algorithm (ITEA)~\cite{ITEA};
15) Deep Symbolic Regression (DSR)~\cite{petersen2019deep};
16) Bayesian Symbolic Regression (BSR)~\cite{BSR}; 
17) Symbolic-Numeric Integrated Pre-training (SNIP)~\cite{SNIP}.

We provide details of the baselines, implementation details, and environmental settings in \textbf{Appendix~\ref{appendix:exp_setting}}.
\vspace{-0.2cm}
\subsection{Overall Performance}
In this experiment, we compare {\model} with the backbone E2E and various SRBench algorithms to evaluate the overall performance. As shown in \textbf{Table~\ref{exp:overall_performance}}, {\model} consistently outperforms E2E in terms of $R^2 > 0.99$ and $R^2$ across the three datasets while maintaining equation complexity that is either lower than or comparable to the backbone. The improvement can be attributed to two key factors:
1) we incorporate symbolic knowledge into the E2E model and fine-tune it for a specific domain, enabling the model to better capture underlying mathematical relationships within the data.
2) By optimizing the latent embedding space, we guide the model to generate more precise and simple equations, enhancing both fitness accuracy and simplicity.
\textbf{Figure~\ref{fig:3in1}} compares {\model} with various SRBench baselines and reveals two notable findings:
1) {\model} surpasses most state-of-the-art baselines in terms of $R^2>0.99$, $R^2$ and inference time.
2) While some methods, such as Operon, achieve a slightly higher proportion of $R^2>0.99$, they have a significantly higher inference time cost, whereas {\model} maintains a competitive $R^2$ while being computationally much more efficient.
Overall, {\model} effectively balances accuracy, equation complexity, and computational efficiency, making it compelling for equation discovery tasks.

\setlength{\tabcolsep}{0.9mm}{
\begin{table}[]
\centering
\small
\caption{
We compared the performance of three fine-tuning strategies: (1) Frozen: we only fine-tune the final layers of the decoder and the whole equation encoder; (2) $\sim$Frozen: we load the pre-train weights but fine-tune all parameters; (3) LoRA: we insert trainable low-rank adaptation modules into the model while freezing the original backbone parameters; (4) Random: we randomly initialize all params. We report the average training time per five epochs.}
\vspace{-0.2cm}
\begin{tabular}{@{}c|c|cccc@{}}
\toprule\toprule
Dataset                    & Method  & $\uparrow R^2$ > 0.99 & $\uparrow R^2$     & $\downarrow$ complexity & $\downarrow$ train time (s) \\ \midrule
\multirow{3}{*}{Feynman}   & frozen  & 0.874  & 0.9833   & 57.76      & 358.8          \\
                           & $\sim$frozen & 0.773      & 0.9442      & 65.95          & 447.5          \\
                           & $LoRA$  & 0.823      & 0.9678       &  63.44           & 400.5           \\       
                           & random  & 0.000      & -2.5076      & 19.09          & 447.1              \\ \midrule
\multirow{3}{*}{Strogatz}  & frozen  & 0.786  & 0.9818 & 50.11      & 31.1           \\
                           & $\sim$frozen & 0.714  & 0.9623 & 53.50      & 39.7           \\
                           & $LoRA$  & 0.714      & 0.9756       &  53.20          & 36.8           \\
                           & random  & 0.000      & -1.7721      & 17.86          & 39.8              \\ \midrule
\multirow{3}{*}{Black-box} & frozen  & -      & 0.7985 & 79.81      & 244.8          \\
                           & $\sim$frozen & -      & 0.7768      & 84.42        & 292.8              \\
                           & $LoRA$  & -      & 0.7798       &  81.33           & 278.6           \\
                           & random  & -      & -5.3426      & 34.77          & 292.4              \\ \bottomrule\bottomrule
\end{tabular}
\vspace{-0.3cm}
\label{exp:transfer}
\end{table}}
\vspace{-0.2cm}
\subsection{Transfer Knowledge Evaluation}
The experimental results in \textbf{Table~\ref{exp:transfer}} demonstrate the critical role of pretraining and fine-tuning strategies in equation discovery. We find that:
1) The frozen approach, which fine-tunes only the final layers of the decoder and the equation encoder, achieves the highest $R^2 > 0.99$ and $R^2$ values across all datasets while maintaining relatively low equation complexity and reduced training time. In contrast, the $\sim$Frozen method, which fine-tunes all parameters while retaining pre-trained weights, results in slightly lower $R^2 > 0.99$ and $R^2$ values and increased equation complexity. The underlying driver is that excessive fine-tuning may lead to overfitting and the generation of unnecessarily complex equations.
2) The LoRA strategy achieves higher performance than ~Frozen while training fewer parameters, but still falls short of the Frozen approach in both accuracy and complexity, likely because the low-rank updates further limit the capacity of an already compact model, making it harder to capture precise symbolic relationships.
3) The Random initialization strategy performs the worst, producing meaningless equations with negative $R^2$ values, highlighting the necessity of pre-trained weights in capturing meaningful mathematical structures. Moreover, while $\sim$Frozen and Random exhibit similar training times, the latter fails to learn valid equations, which indicates the importance of knowledge transfer from pretraining. These results indicate that Frozen fine-tuning provides the best balance between fitness accuracy, efficiency, and simplicity, effectively leveraging pre-trained mathematical priors while avoiding unnecessary computational overhead.
These results suggest that Frozen fine-tuning provides the best balance between fitness accuracy, efficiency, and simplicity, effectively leveraging pre-trained mathematical priors while avoiding unnecessary computational overhead.

\begin{figure}[t]
  \centering
  \subfigure[\small{Feynman}]{
    \includegraphics[width=0.227\textwidth]{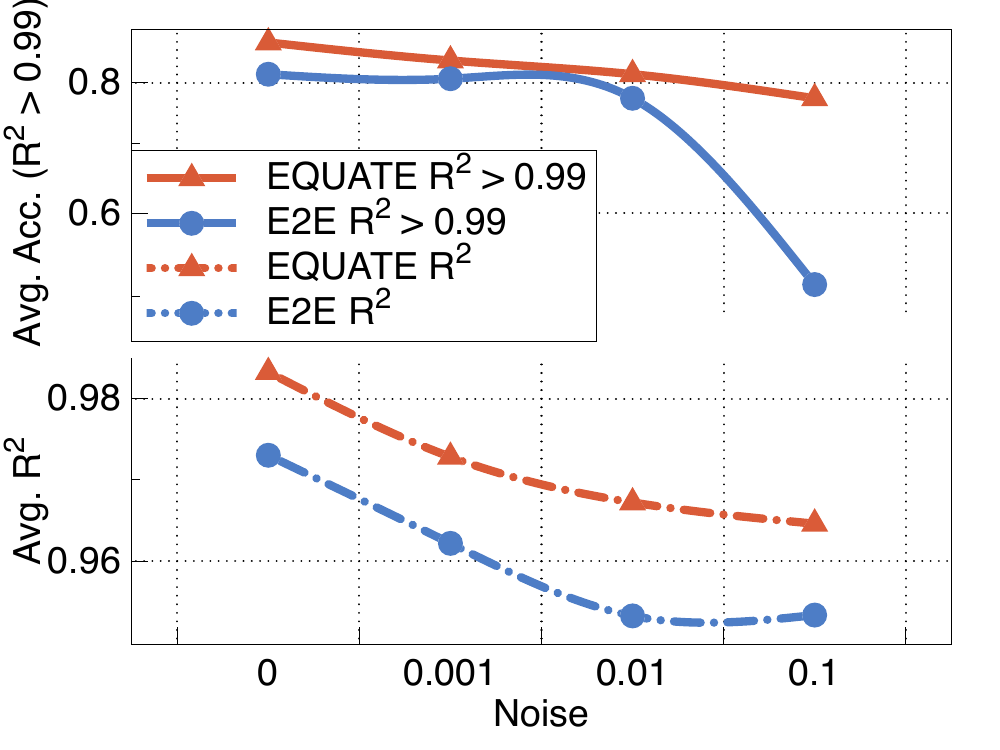}
  }
  \hfill
  \subfigure[\small{Strogatz}]{
    \includegraphics[width=0.227\textwidth]{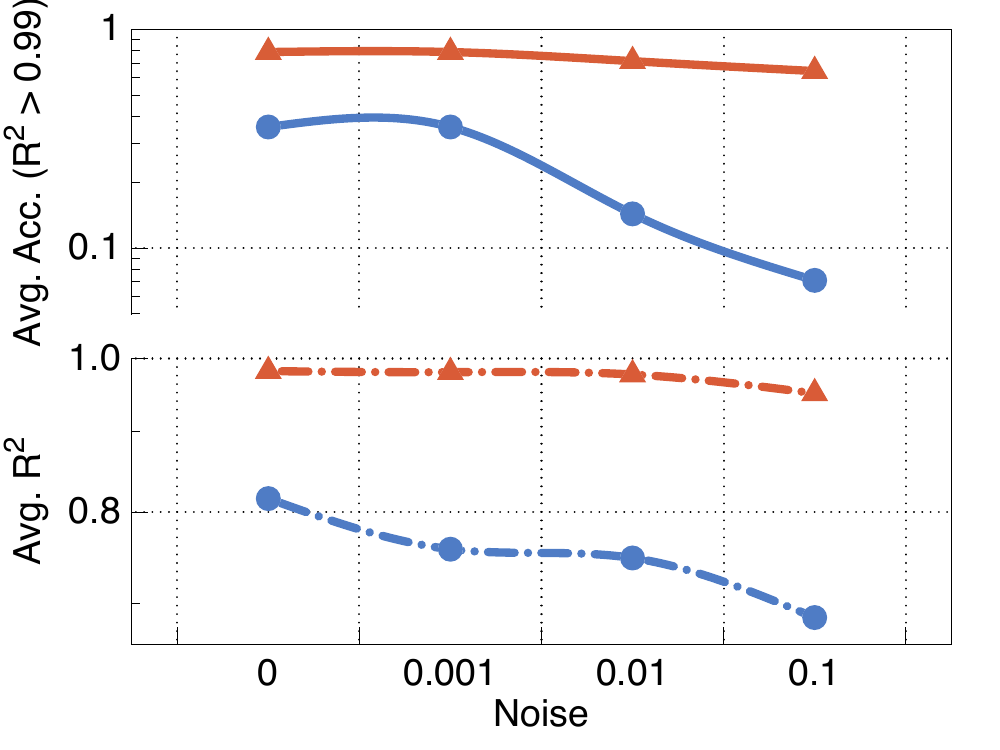}
  }
  \vspace{-0.5cm}
  \caption{Performance in terms of robustness to noise.}
  \vspace{-0.5cm}
  \label{fig:robust}
\end{figure}
\vspace{-0.2cm}
\subsection{Robustness to Noise in Transfer Tasks}
We inject noise into the golden labels $y$ of the training set to evaluate the {\model}'s robustness. \textbf{Figure~\ref{fig:robust}} illustrates the robustness of {\model} and E2E methods under increasing noise levels. As noise increases, both models experience a decline in performance, but the drop is significantly steeper for E2E, particularly in the $R^2 > 0.99$. In contrast, {\model} consistently outperforms E2E across all noise levels, maintaining a higher $R^2$ and greater accuracy, and the performance gap between them widens as noise increases, indicating {\model}’s superior robustness. 1) dataset-specific differences reveal that in the Feynman dataset, both models gradually degrade, whereas in the Strogatz dataset, {\model} maintains near-constant high performance while E2E suffers a sharp drop, suggesting that E2E is more vulnerable in certain cases. 2) Symbolic priors enable {\model} to generalize better and resist overfitting to noise, while E2E, being purely data-driven, tends to memorize noise perturbations. 3) {\model} employs a differentiable search mechanism and implicit regularization, preventing overfitting to noisy fluctuations, whereas E2E relies on direct function approximation without such safeguards. 4) Pre-trained knowledge transfer further enhances {\model}’s resilience in small-data, high-noise settings, where E2E struggles due to its dependence on numerical regression. These findings highlight the advantages of integrating symbolic priors and structured optimization strategies to improve the robustness of equation discovery models.

\begin{figure}[t]
  \centering
  \subfigure[\small{Feynman}]{
    \includegraphics[width=0.227\textwidth]{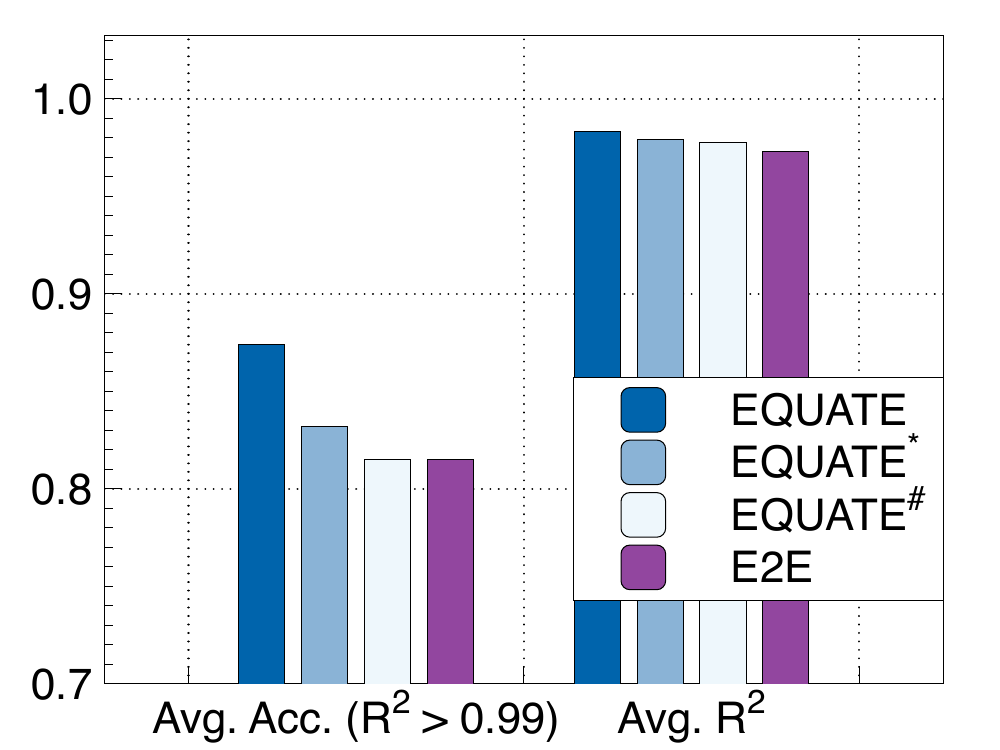}
  }
  \hfill
  \subfigure[\small{Strogatz}]{
    \includegraphics[width=0.227\textwidth]{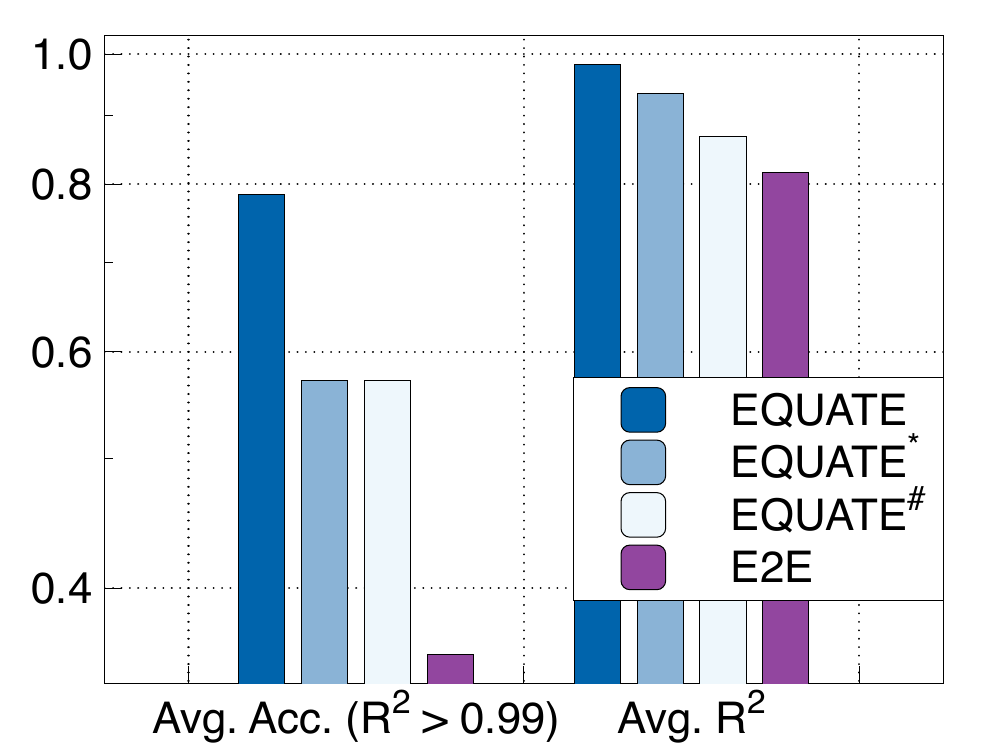}
  }
  \vspace{-0.5cm}
  \caption{Verify the key components of the equation encoder and embedding space optimization.}
  \vspace{-0.5cm}
  \label{fig:ablation}
\end{figure}
\vspace{-0.2cm}
\subsection{Ablation Study}
This experiment aims to validate the effectiveness of two key components: equation encoder (i.e., symbolic knowledge infusion) and embedding space optimization. To achieve this, we performed ablation studies on the Feynman and Strogatz data sets using three model variants:
1) {\model}$^*$, which removes embedding space optimization but retains the equation encoder for symbolic knowledge infusion; 
2) {\model}$^\#$, which removes the equation encoder while fine-tuning the backbone and maintaining embedding space optimization; 
3) E2E, an end-to-end baseline model. 
\textbf{Figure~\ref{fig:ablation}} shows that: 
1) {\model}$^*$ performs slightly worse than {\model} but better than {\model}$^\#$, suggesting that symbolic knowledge infusion plays a crucial role in enhancing generalization and accuracy, especially in scientific domains where structured priors are important. 
2) While {\model}$^\#$ achieves a comparable average $R^2$ to {\model}, but underperforms in high-accuracy cases ($R^2 > 0.99$).This indicates that embedding space optimization improves fit quality, but cannot fully compensate for the absence of explicit symbolic structure. 
3) E2E performs the worst, particularly on the Strogatz dataset, emphasizing the limitations of purely end-to-end training in capturing symbolic structure and learning semantically aligned representations. 
These findings confirm that both symbolic knowledge infusion and embedding space optimization are essential for achieving superior model performance. The symbolic knowledge infusion enhances generalization by providing structured information, and embedding space optimization refines the learned representations for improved effectiveness. To make the experiments more convincing, we analyze the impact of symbolic alignment and gradient search, accuracy-simplicity trade-off through $\lambda$, and complexity check in ~\textbf{Appendix~\ref{appendix:gradient},~\ref{appendix:lambda},~\ref{appendix:complexity}}, respectively.
\vspace{-0.3cm}
\section{Related Work}
Symbolic Regression (SR) aims to derive optimal mathematical expressions for a function based on its observed values. Genetic Programming (GP)~\cite{GP,GP2,GP3,GP4,GP5,GP6,GP7} is a classical SR method that iteratively refines expressions using crossover and mutation. However, GP often suffers from high search complexity, making it computationally expensive and prone to local optima. To improve efficiency, research has integrated GP with Monte Carlo Tree Search (MCTS)~\cite{sun2022symbolic} and reinforcement learning~\cite{mundhenk2021symbolic}, significantly accelerating the search process. Some approaches, such as AI Feynman~\cite{udrescu2020ai}, introduce domain knowledge (e.g., physical laws) to guide equation discovery. However, these methods lack effective mechanisms to transfer prior knowledge across datasets, often restarting the search process from scratch, which limits their generalization and scalability.
The advent of deep learning has introduced neural networks into SR~\cite{petersen2019deep,kim2020integration,zhang2023deep,d2022deep,ying2025bridging}, improving expression generation. Hybrid methods combining GP and neural networks have been proposed to predict expression structures~\cite{cranmer2020discovering}, reducing search difficulty. More recently, Transformer-based generative models have further advanced SR by leveraging large-scale pretraining. For e.g., \cite{biggio2021neural} proposed a Transformer pre-trained on symbolic regression datasets, generating equation skeletons with placeholders for constants. Similarly, \cite{kamienny2022end} introduced an end-to-end symbolic regression framework based on sequence generation. While these models offer improvements in scalability and efficiency, their training objectives—typically borrowed from natural language generation—prioritize token-level similarity, which is misaligned with the core goals of symbolic regression, such as accuracy, interpretability, and parsimony. To mitigate this, decoding strategies such as beam search~\cite{graves2012sequence,wiseman2016sequence} and sampling~\cite{fan2018hierarchical} provide multiple candidate equations for optimization, but their effectiveness depends heavily on the accuracy of the decoding process. MCTS-based decoding~\cite{shojaee2023transformer} has been proposed to enhance symbolic reasoning but remains computationally intensive and highly dependent on the quality of pretraining. Nevertheless, the computational cost and dependency on the quality of pre-training data remain significant challenges. Recent work has also shown that large language models (LLMs) can perform symbolic regression in a zero-shot setting by leveraging pretraining on mathematical corpora~\cite{shojaee2024llm}. However, such models often struggle with small, domain-specific datasets due to a lack of adaptation and a tendency toward negative transfer. These limitations highlight the importance of fine-tuning strategies, such as our proposed EQUATE framework, which adapts foundation models for domain-specific symbolic equation discovery by integrating symbolic priors and data-fit supervision in a structured embedding space.
By combining feature engineering with generative AI, this work enables robust equation discovery that extends beyond methodological advances to practical impact~\cite{ying2023self,ying2024feature,ying2024unsupervised,ying2024topology,gong2025neuro}. It can support sustainability in renewable energy forecasting~\cite{huo2025ct}, strengthen financial security through fraud detection~\cite{huo2025enhancing}, enhance creative image applications~\cite{zhu2025image}, and further benefit areas such as simulation~\cite{bai2025brownian}, bioinformatics~\cite {ying2024revolutionizing}, and material science~\cite{hu2024reinforcement}.
\section{Limitation}
The current \model\ is constrained by its end-to-end (E2E) encoder-decoder architecture. Due to efficiency and memory constraints, the model limits input dimensionality to 10, which restricts its applicability to high-dimensional datasets. This restriction is necessary to avoid excessively long token sequences after transformation, which would otherwise impair training and inference efficiency. In future work, we plan to investigate more scalable architectures or hybrid encoding strategies that can better handle high-dimensional inputs while maintaining efficiency.

\section{Conclusion}
In this work, we propose \model\ (Equation Generation via Quality-Aligned Transfer Embeddings), a fine-tuning framework that effectively adapts foundation model knowledge to domain-specific equation discovery through symbolic-numeric alignment and evaluator-guided optimization.
Our key contributions include: (i) Lightweight Fine-Tuning for Symbolic Regression: A framework that adapts pre-trained foundation models to symbolic equation generation using only small, domain-specific datasets, mitigating negative transfer. (ii) Joint Symbolic-Numeric Representation: A dual-encoder architecture that aligns symbolic equations and numerical data in a shared embedding space, enhancing integration of domain knowledge and observed patterns. (iii) Evaluator-Guided Embedding Optimization: A differentiable evaluator scores equation fitness and simplicity, enabling structured, domain-aware search beyond standard token-level decoding. (iv) Fitness-Based Symbolic Generation: \model\ leverages fitness supervision during both training and inference to generate more accurate and parsimonious equations, particularly under data-scarce conditions. Our experiments show that \model\ significantly improves equation discovery accuracy, lowers computational cost, and enhances robustness to noisy data. These results demonstrate the practical value of \model\ for scientific discovery and AI-driven symbolic regression, providing an effective solution for knowledge transfer in equation discovery.
\bibliographystyle{ACM-Reference-Format}
\bibliography{main_content/ref}
\appendix
\section{Experimental Settings}
\label{appendix:exp_setting}
\noindent\textbf{Implementation Details.}
We leverage the state-of-the-art open-source End-to-End (E2E) model~\cite{kamienny2022end} as the pretrained transformer backbone. This selection is due to the public availability of E2E's model architecture, pretrained weights, and output logits through the Facebook symbolic regression library and its associated repository\footnote{\url{https://github.com/facebookresearch/symbolicregression}}.
\ul{1) Numerical Encoder and Decoder:} We reload the E2E pre-trained model and ensure the hyperparameters of these two components remain consistent with~\cite{kamienny2022end}.
\ul{2) Symbolic Encoder:} A single-layer LSTM encodes symbolic sequences into 512-dimensional embeddings with a maximum token sequence length of 200.
\ul{3) Evaluator:} We implement the evaluator using a two-layer MLP with a hidden dimension of 1024.
\ul{4) Data Generation:} For each test equation, we perform replacement sampling to generate 100 observation sets from the training data, with each set containing 200 data points. To enhance data diversity and quality, 10\% of the equations are randomly generated, while 90\% were created using the backbone model as symbolic knowledge.
\ul{5) Fine-tuning:} We fine-tune the model for 10 epochs with a learning rate of 1e-4 using the AdamW optimizer, freezing the last transformer layers of the decoder. The scaling factors $\alpha=0.05$, and $\beta=100$ align the reconstruction loss and MSE loss on the same scale.
\ul{6) Optimization:} To maintain consistency with the backbone (a maximum of 10 bags), we set the number of initial points to 10, the search step size to a maximum of 20 steps, and the total number of generated equations to no more than 100. The search process stops immediately if an equation with $R^2 > 0.99$ is found.

\noindent\textbf{Baselines.} We compare {\model} with the backbone E2E and various SRBench algorithms\footnote{\url{https://github.com/cavalab/srbench}} to evaluate our proposed method.
\textbf{1) End-to-end symbolic regression with transformers (E2E):} E2E~\cite{kamienny2022end} trains Transformers to directly generate full symbolic expressions, including constants, from input data in a single pass, offering high accuracy and orders-of-magnitude faster inference than GP-based methods.
\textbf{2) Operon:} Operon~\cite{GP} is a fast C++ genetic programming framework for symbolic regression with linear tree encoding and parallel evolution.
\textbf{3) Semantic Backpropagation GP (SBP-GP):} SBP-GP~\cite{SBP-GP} uses semantic backpropagation to guide subtree replacement in genetic programming, and is enhanced with linear scaling to improve symbolic regression accuracy and generalization.
\textbf{4) Transformer-based Planning for Symbolic Regression (TPSR):} TPSR~\cite{shojaee2023transformer} integrates Monte Carlo Tree Search with pretrained Transformer SR models to guide equation generation using feedback on accuracy and complexity, improving both fitting and interpretability.
\textbf{5) GP-based Gene-pool Optimal Mixing Evolutionary Algorithm (GP-GOMEA):} GP-GOMEA~\cite{GP-GOMEA} applies gene-pool optimal mixing to genetic programming, using linkage learning and entropy-based building-block identification to improve scalability and solution compactness.
\textbf{6) Epsilon-Lexicase Selection (EPLEX):} EPLEX~\cite{EPLEX} is a parent selection method for symbolic regression that relaxes elitism in lexicase selection using adaptive $\epsilon$ thresholds, improving performance in continuous domains.
\textbf{7) Multiple Regression GP (MRGP):} MRGP~\cite{MRGP} enhances symbolic regression by linearly combining program subexpressions via multiple regression, using them as features to improve predictive accuracy.
\textbf{8) Age-Fitness Pareto Optimization (AFP):} AFP~\cite{AFP} maintains evolutionary diversity by selecting individuals on a Pareto front of fitness and age, helping avoid premature convergence in symbolic regression.
\textbf{9) Age-Fitness Pareto Optimization with Co-evolved Fitness Predictors 
 (AFP\_FE):} AFP\_FP~\cite{AFP} extends AFP optimization by co-evolving fitness predictors to accelerate evaluation, enabling efficient symbolic regression on large datasets.
\textbf{10) AIFeynman:} AIFeynman~\cite{udrescu2020aifeynman20paretooptimal} combines neural networks, graph modularity discovery, and Pareto-optimal pruning to robustly recover interpretable symbolic equations from noisy data.
\textbf{11) Feature Engineering Automation Tool (FEAT):} FEAT~\cite{FEAT} evolves networks of syntax trees with differentiable weights to learn compact and interpretable representations for symbolic regression.
\textbf{12) gplearn:} gplearn performs symbolic regression using genetic programming, providing interpretable models through a scikit-learn-style interface.
\textbf{13) Fast Function Extraction (FFX):} FFX~\cite{FFX} is a deterministic symbolic regression method that combines massive basis function enumeration with pathwise regularized learning to efficiently extract compact, interpretable models.
\textbf{14) Interaction-Transformation Evolutionary Algorithm (ITEA):} ITEA~\cite{ITEA} evolves structured affine combinations of nonlinear feature interactions using mutation-only genetic search, enabling interpretable and scalable symbolic regression.
\textbf{15) Deep Symbolic Regression (DSR):} DSR~\cite{petersen2019deep} trains an autoregressive RNN with a risk-seeking policy gradient to generate symbolic expressions that maximize best-case performance in symbolic regression.
\textbf{16) Bayesian Symbolic Regression (BSR):} BSR~\cite{BSR} frames symbolic regression in a Bayesian framework by linearly combining symbolic trees sampled via MCMC, enabling interpretable expressions and incorporation of prior knowledge.
\textbf{17) Symbolic-Numeric Integrated Pre-training (SNIP):} SNIP~\cite{SNIP} combines symbolic and numeric information via contrastive pretraining to construct a foundation model.

\noindent\textbf{Environmental Settings.}
All experiments are conducted on the Ubuntu 22.04.3 LTS operating system, Intel(R) Xeon(R) w9-3475X CPU@ 4800MHz, and 1 way RTX A6000 and 48GB of RAM, with the framework of Python 3.11.4 and PyTorch 2.5.1.

\begin{figure}[h]
  \centering
  \subfigure[\small{strogatz\_vdp1}]{
    \includegraphics[width=0.227\textwidth]{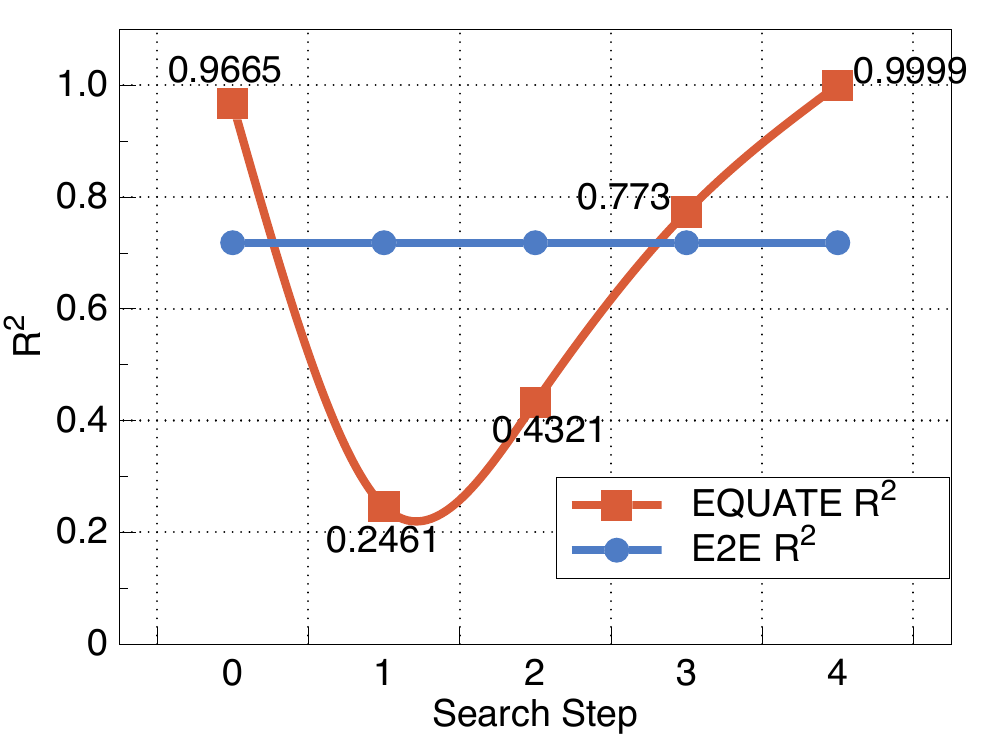}
  }
  \hfill
  \subfigure[\small{Strogatz}]{
    \includegraphics[width=0.227\textwidth]{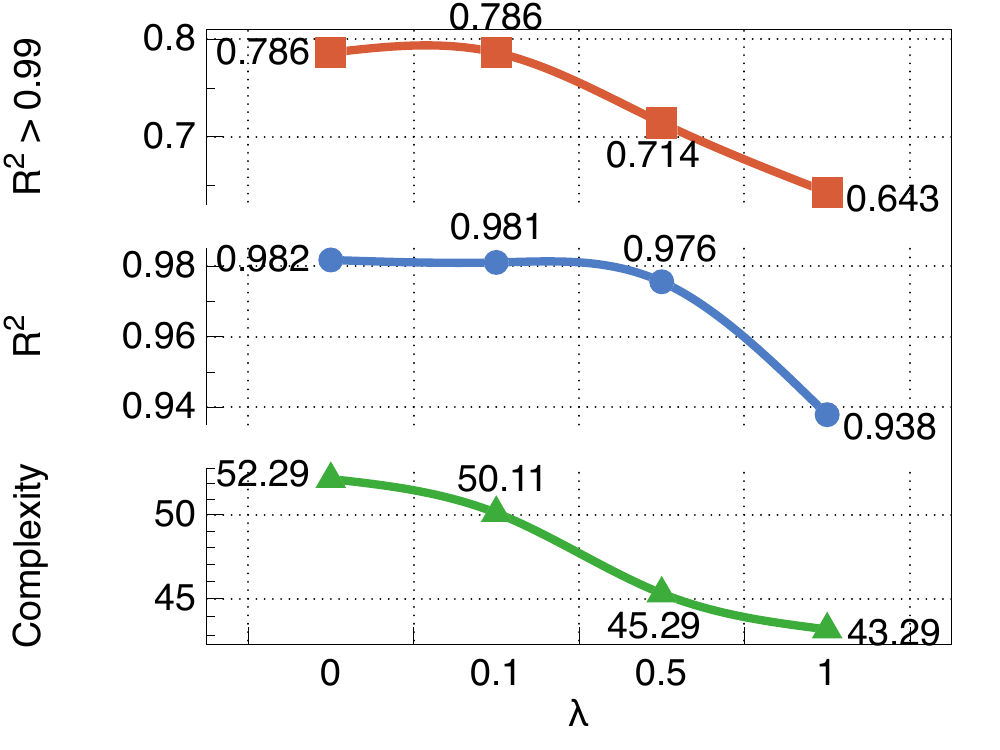}
  }
  \caption{Performance in terms of gradient search (a) and the $\lambda$ trade-off (b).}
  \label{fig:extra}
\end{figure}

\section{Impact of Symbolic Initialization and Gradient Refinement}
\label{appendix:gradient}
To assess the impact of the gradient search mechanism in \model, we evaluate both \model~and the E2E baseline on the same data matrix derived from the strogatz\_vdp1 function. We track performance across multiple search steps, with step 0 representing the initial embedding and subsequent steps representing refinement guided by the evaluator.
As shown in the \textbf{Figure~\ref{fig:extra}(a)}, \model\ achieves a significantly higher initial performance at step 0 (0.9665) than the E2E baseline (0.718), demonstrating the benefit of symbolic-numeric alignment in initializing latent representations. Furthermore, with evaluator-guided gradient search, the \model exhibits performance oscillation during intermediate steps, but ultimately converges to near-perfect accuracy (0.9999) at step 4.
These results highlight the advantage of combining symbolic priors with gradient-based refinement. \model\ not only starts from a better initialization but also consistently improves through feedback-driven embedding updates, ultimately outperforming the end-to-end model.

\section{Exploring the Accuracy-Simplicity Trade-off via the $\lambda$ Parameter}
\label{appendix:lambda}
To investigate the trade-off between accuracy and functional simplicity, we conduct controlled experiments on the Strogatz dataset by varying the trade-off parameter $\lambda$ in the evaluator objective. \textbf{Figure~\ref{fig:extra}(b)} shows that the parameter $\lambda$ balances the weight between accuracy and simplicity during model evaluation and optimization.
As $\lambda$ increases, the evaluator prioritizes simplicity over accuracy. This shift results in recovered functions that are structurally simpler (lower complexity scores), but with a noticeable drop in prediction performance. For instance, moving from $\lambda = 0$ to $\lambda = 1$ reduces complexity from 52.29 to 43.29, while the proportion of accurate predictions drops from 0.786 to 0.643. These results illustrate the expected trade-off and highlight the need to tune $\lambda$ depending on application goals.

\begin{figure}[h]
  \centering
  \subfigure[\small{Training Time}]{
    \includegraphics[width=0.227\textwidth]{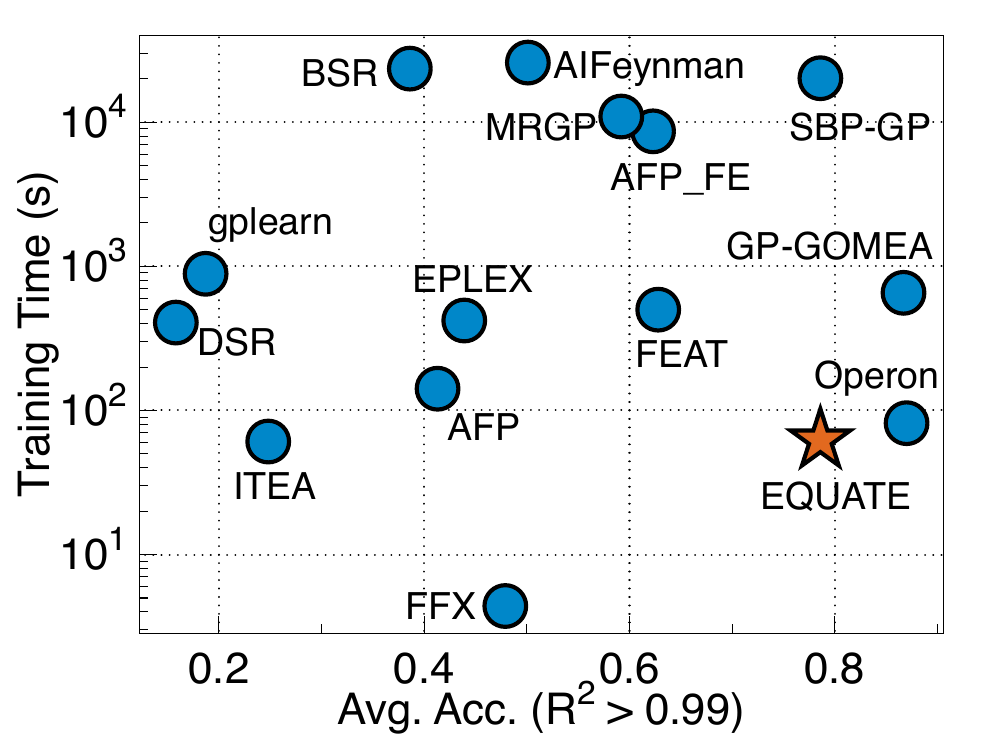}
  }
  \hfill
  \subfigure[\small{Model Size}]{
    \includegraphics[width=0.227\textwidth]{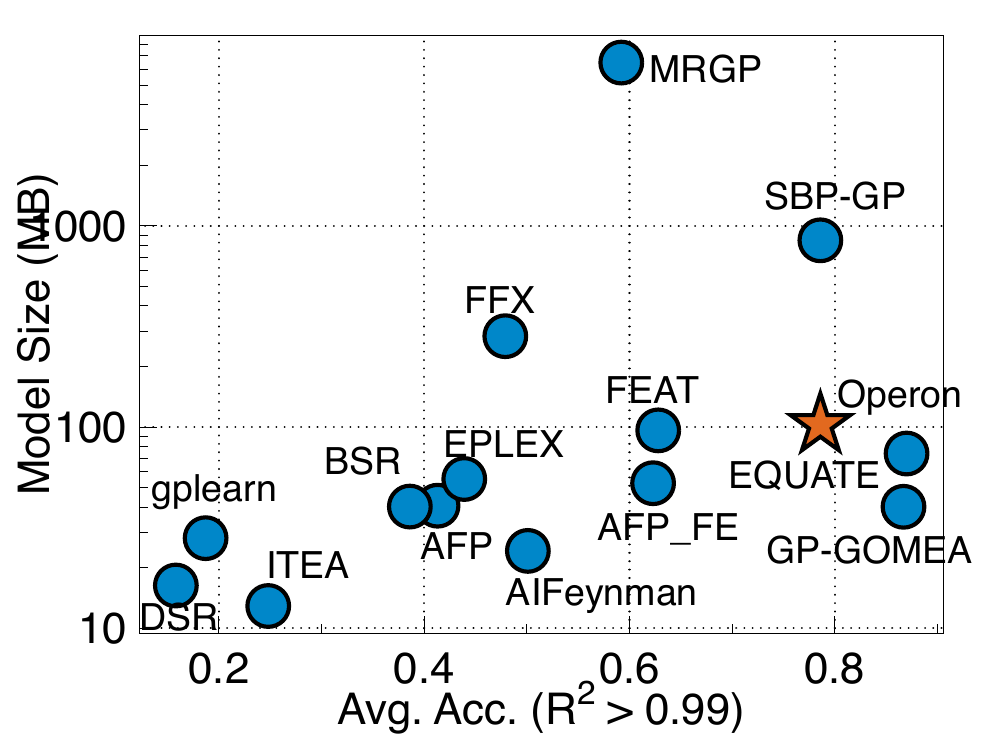}
  }
  \caption{Complexity check on Strogatz dataset. {\model} exists strong accuracy-speed-model size tradeoffs.}
  \label{fig:complexity}
\end{figure}
\section{Complexity Check}
\label{appendix:complexity}
This experiment aims to evaluate the trade-offs between accuracy, training time, and model size between {\model} and various baselines on the Strogatz dataset.
1) \textbf{Figure~\ref{fig:complexity}(a)} shows that BSR, MRGP, and AIFeynman require extensive computational resources, yet achieve lower accuracy. FFX trains significantly faster but still struggles with accuracy. {\model} and Operon achieve high accuracy while maintaining shorter training times. However, it should be noted that once the model converges, the inference time of {\model} is significantly lower than that of Operon, as shown in \textbf{Figure~\ref{fig:3in1}}. This highlights the efficient fine-tuning mechanism of {\model}, which effectively reduces computational overhead while preserving accuracy.
2) \textbf{Figure~\ref{fig:complexity}(b)} examines model size versus accuracy, revealing that high-accuracy models such as SBP-GP require substantial storage, whereas lower-accuracy models (e.g., ITEA, DSR, gplearn) sacrifice accuracy for compactness. {\model} achieves a favorable trade-off, which delivers high accuracy while maintaining a more compact model size than most baselines.
These findings indicate that most baselines struggle to optimize all three factors simultaneously, while {\model} effectively balances accuracy, efficiency, and model compactness, making it competitive for equation discovery tasks.
\end{document}